# Hidden Structure in the Solution Set of the N-Queens Problem

T. E. Raptis[abc]

[a]National Center for Science and Research "Demokritos", Division of Applied Technologies, Computational Applications Group, Athens, Greece.

[b]University of Athens, Department of Chemistry, Laboratory of Physical Chemistry, Athens, Greece

[c]University of Peloponnese, Informatics and Telecommunications Dept., Tripoli, Greece.

E-mail: traptis@protonmail.com

**Abstract:** Some preliminary results are reported on the equivalence of any n-queens problem with the roots of a Boolean valued quadratic form via a generic dimensional reduction scheme. It is then proven that the solutions set is encoded in the entries of a special matrix. Further examination reveals a direct association with pointwise Boolean fractal operators applied on certain integer sequences associated with this matrix suggesting the presence of an underlying special geometry of the solutions set.

## 1. Introduction

Recently, two major problems in complexity theory were linked via a set of mappings, indicating a deeper unity that deserves further examination. In [1], a surprisingly simple association between the 2D Ising model and staisfiability (SAT) problems was proven while, in [2] the N-Queens problem was also linked to certain instances of the SAT problem with a rather pessimistic conclusion for general A.I. Lately, two new studies appeared on the possibility of a direct quantum implementation of the same problem either as a standard quantum gates circuit [3], or as an Ising spin type of machine [4], these later evolutions pointing towards possible tests of the acclaimed quantum supremacy.

From the point of view of computer science and abstract mathematics, the particular problem belongs to the general class of constraint satisfaction problems [5] while in terms of set theory it is known to be a generalization of exact cover problems which include pentomino tilings and Sudoku solvers [6]. Knuth, was able to accelerate search strategies for such problems with his "*X*" algorithm in [7]. A variety of classical algorithms have been tried which are reviewed in [8] while some greedy algorithms [9] are capable of true $O(N^a)$ with *a* of the order of 2 or 3 at the cost of locating only partial solutions. Several open questions exist for this problem including the absence of any analytical partition function for counting all solutions and even asymptotic expressions are lacking.

In the present work, we analyze the N-Queens problem from a different perspective introducing a different approach based on a kind of dimensional reduction to the complete dictionary of binary words of $N^2$ length as the equivalent $2^{N^2}$ search space and a special type of connectivity matrix which was termed, the "*Interaction Kernel*", a term which we shall keep here for convenience. The particular technique had been previously introduced in the context of Cellular Automata (CA) and their global maps [10] for arbitrary lattice dimensionality while in the present application we only need to deal with the 2D case of a standard chessboard.

Natural extensions of the problem in higher dimensions do exist [11] but they will not be dealt with in the present report although the same strategy is expected to work at the cost of higher complexity kernels. We also notice that one can define "modular" N-queen problems with toroidal boundary conditions with the sole difference being that the x-shaped parts of any neighborhood become periodic trajectories always covering the chessboard in a transitive manner. For the rest of the paper, we restrict attention to the case of fixed boundaries.

In the next section, the exact method of construction of the relevant kernel matrix is given together with some linear algebraic properties while in section 3, a set of equivalent conditions to be satisfied by any solution is presented. In section 4 we redefine the problem via an additional transformation of these conditions which makes contact with some well known arithmetic fractals and the solutions set is restricted to a special set of paths in a special function defined on an N-dimensional hypercube.

## 2. The kernel method for dimensional reduction

While previous attempts for solving the n-queens using backtracking or similar techniques retained the 2D character of the chessboard array we can bypass this difficulty based on very general arguments already laid out in [10]. The essential difference from the CA case lies in the topology of the neighborhood for which reason we will have to use a different kind of mapping after reduction. We perform the reduction via a reshaping of the original $N$ x $N$ chessboard matrix into a $N^2$ length vector formed by a sequence of the original matrix rows. We may denote such an operation over any $n$ x $m$ matrix $M$ as vec$\{M\}$ = $[M_{11},…,M_{1n}, M_{21},…,M_{2n},…,M_{N1},…,M_{nm}]$. In array languages like Matlab, Octave or Scilab, this option is given via a "*reshape*" command.

We then need to establish an auxiliary connectivity matrix into which the topology of any neighborhood of possible interactions between individual entries of the original chessboard matrix will be preserved. In the case of standard CAs this comes in the form of so called, Von Neumann and Moore types of nearest neighbor areas [12]. In the case of the N-queen problem there are certain types of constraints regarding rows and columns containing exactly one pawn and diagonals that are allowed to contain at most one or none.

To each queen we assign an open neighborhood restricted only by the board's boundaries or the queen's neighborhood range via a multiplexing scheme using an appropriate auxiliary traceless, symmetric Boolean matrix $K$. This will stand for the "Interaction Kernel" where all accessible positions around each queen's central position corresponding to its the diagonal are marked with ones the rest of the positions being marked with zeros and ignored. Since any such neighborhood in the original 2D array is formed as a superposition of "crosses" and x-shaped rays, it is straightforward to form sequentially all logical rows of $K$ with the relevant accessible positions by moving each central position and reshaping. The resulting matrix is symmetric and has an almost circulant structure as shown in figure 1 for an ordinary 8 x 8 chessboard. In figure 2, we show a representative of the eigenvalue spectrum of the Boolean kernel with all eigenvalues real.

Construction of the kernel can be simplified using tensor product algebra via the two main constituents given as $\mathbf{K} = \mathbf{K}_+ + \mathbf{K}_\times$ where the "+" and "x" indices represent the superposition of the two constituents of each queen's shooting range being composed independently from rows and columns as well as diagonals and anti-diagonals respectively. We then examine the structure of each of the sub-kernels separately. To this aim we shall also need to introduce the Boolean complement of an identity matrix as $\bar{\mathbf{I}}_L = \mathbf{1}_L - \mathbf{I}_L$ where the index $L$ denotes an $L$ x $L$ block and $\mathbf{1}$ denotes an all ones block.

For the case of row-column kernel it is possible to factorize and also use the Kronecker sum rule as

$$\mathbf{K}_+ = \mathbf{I}_L \otimes \bar{\mathbf{I}}_L + \bar{\mathbf{I}}_L \otimes \mathbf{I}_L = \bar{\mathbf{I}}_L \oplus \bar{\mathbf{I}}_L$$

(1)

The case of the second diagonal part of the kernel is much less trivial and requires a special expansion of the diagonal structure of any $[\mathbf{1}_L]$ block in the form

$$\mathbf{1}_L = \sum_{i=1}^{L-1} \mathbf{k}_i^L$$

(2)

In (2) we use a decimating set of $k$ matrices with a double pair of unique diagonals and anti-diagonals making a set of discrete parallelograms of the form

$$\mathbf{k}_1^L = \begin{bmatrix} 0 & \cdots & 1 \\ 0 & 1 & 0 \\ 1 & \cdots & 0 \end{bmatrix}, \quad \mathbf{k}_2^L = \mathbf{I}_2 \otimes \mathbf{J}_{L^2-2} = \begin{bmatrix} 0 & 0\cdots & 1 & 0 \\ 0 & 1 & 0 & 1 \\ 1 & 0 & 1 & 0 \\ 0 & 1 & 0\cdots & 0 \end{bmatrix}, \ldots$$

where $\mathbf{J}_i$ is the i x i unit anti-diagonal. This makes possible the full decomposition of the second part of the kernel as

$$\mathbf{K}_\times = \sum_{i=1}^{L-1} (\mathbf{k}_i^L \otimes \mathbf{k}_i^L)$$

(3)

We notice that none of the kernel's rows cannot belong to any solution set by construction since all positions of any neighborhood have been filled apart from the central one thus having a bit summand much higher than any $S$ vector. In figure 3(a), we also present the row-wise or column-wise bit summands while in 3(b) we present the evolution of upper and lower bounds of the same with the board size. In the next section we analyze the action of the kernel and extract the most important property of the kernel rows and its direct link with possible solution vectors.

## 3. Decoding the solution set from the Boolean kernel

We first examine the action of the kernel on any flattened configuration vector representing an arbitrary checkerboard. For any $L \times L$ board, there will in general exist $2^{L^2}$ possible Boolean configuration vectors for any queens arrangement with $2 < N \leq L$. These can be subdivided into exactly $L$ distinct binomial sets $P_N$, $N=2,\ldots,L$, of pattern permutations with cardinality $\#(P_N) = \binom{L^2}{N}$ each. Given an integer encoding $v$ of each such vector via the polynomial representation in a lexicographically ordered dictionary over any exponential interval, one can also associate $N$ with $s_2(v) = N$ where we identify $s_2$ as the binary digit-sum function [15] of which direct evaluation is possible as $\sum_{i=0}^{L^2-1} [2^{-i} v](\mod 2)$. A direct serial search over each $P_N$ would require $\#(P_N)$ steps for an exhaustive examination of each pattern. Any faster method must reduce the total search space.

Assume then, a particular instance of a configuration vector $\vec{S}^{v \in P_N}$ with exactly $N$ 1s and an ordered set of integer bit pointers serially marking any particular instance of combinations inside $P_N$ as $p = [p_1,...,p_N]$ with a standard $C$ convention (counting from zero) such that the corresponding integer index would be of the form $v = \sum_{j=1}^{N} 2^{p_i}$. Assume also a subset $\Sigma_N \subset P_N$ such that $\vec{S}^{v \in \Sigma_N}$ represents a valid N-queens configuration. Then the vector $\mathbf{K} \cdot \vec{S}^{v \in \Sigma_N}$ will immediately map to zero all positions corresponding to the rows $K_{\{pi\}}$ with the rest containing arbitrary overlaps with irrelevant positions. To extract an indicator we use an orthogonal complement leading to a quadratic as

$$\vec{S}^{\,T} \cdot \mathbf{K} \cdot \vec{S} = 0 \qquad (4)$$

The above while true for all valid configurations is a crude binary classifier that cannot be used to totally discriminate the particular subset. To make a perfect classifier one can instead use a mapping of the form

$$\forall i, j : K_{ij} = 1 \rightarrow Q_{ij} = 2^j, \ Tr(\mathbf{Q}) = 0 \qquad (5)$$

It is evident that the new kernel $Q$ will project the exact solution set for all $P_N$ to a power of 2. The reasoning is simply that the polynomial representation of each neighborhood acts as a kind of restricted Gödel code for all the possible classes of any bit pattern in any $S$ vector thus making a 1-1 mapping of any neighborhood to some unique integer with a perfect power of 2 only for a valid configuration.

Methods like the above still have a total complexity equal to that of a single matrix-vector product, generally between $O(N^2)$ and $O(N^3)$ multiplied with the associated binomial coefficient for each set $P_N$. The latter can be constructed efficiently with backtracking methods based on the fact that all bit patterns of each set admit a lexicographic Gray code [16], [17]. Our purpose in what follows is different than reducing the complexity. It is possible to show that at least the highest order solution set $P_L$ can be located via the sole examination of the Boolean kernel contents.

The critical observation from (4) starts with the recognition that at least for any valid subset $\Sigma_N$ those rows of $K$ that project to zero must have complementary bits with the corresponding configuration vector $\vec{S}^{v \in \Sigma_N}$. As a matter of fact, for every non-empty position of any such vector, the corresponding counters vector $p$ gives the equivalent $K$ rows such that the total subset of rows must have a common minimal subset of zero positions reflecting the exact structure of the common configuration vector buried in other noisy bits from adjacent neighborhoods. Isolating this particular subset reveals the structure of a special rank one dyadic matrix as $SS^T$ inside the kernel. We may formalize this equivalence using the following property.

Assume a state vector $\vec{S}^{v \in \Sigma_N}$ and the associated p-vector marking the positions of 1s or valid queen's positions and using $\vec{K}_i$ to denote a particular kernel row as a vector, the following should hold

$$\vec{S}^{v \in \Sigma_N} = \prod_{i \in \vec{p}}^{L} \vec{K}_i \tag{6}$$

where we interpret the product as a pointwise one between row elements. An example of the dyadic structure is given in figure 4(a,b), with the known fundamental solutions of both the *6 x 6*, *N=6* problem with a p-vector as [4, 7, 17, 20, 30, 33] and the *7 x 7*, *N=7* problem with a p-vector as [2, 11, 15, 28, 33, 38, 48]. Property (6) can be verified directly by explicitly constructing each kernel. A simple Matlab code for a kernel constructor is given in appendix A.

Apart from the fundamental there are more solutions out of symmetry operations. Since by construction, the kernel must be compatible with all possible solutions there must a unique maximal closure of all such hidden dyadic submatrices. In general the maximal closure of all dyadics will contain every possible superposition. This can always be thought as a matrix summand in which case it could be separated into several simpler problems where one is asked to recognize the presence of a unique dyadic submatrix inside the main kernel matrix. We thus arrive at a different formulation of the original problem of equal complexity, the problem now being that of isolating the maximal closure for all rank one submatrices of dyadic form from the main kernel matrix. In the next section we treat this in connection with some other well known combinatorial problems and further analyze the underlying structure.

4. **Subgraph isomorphism and Boolean fractals**

The general case of identifying an arbitrary submatrix inside another, also appears in the literature as having a direct correspondence with the well known, subgraph isomorphism problem [18], [19] which is provably NP-complete. Certain special classes of this problem have already been proven to admit polynomial time solutions [20], [21] while the matrix representation of generic graphs has been used in a recently introduced Ordered Matrix Matching (OMM) method for security applications [22]. In the language of general signal and image processing, such algorithms can also be interpreted as special cases of "matched filters" [23] that generalize correlation based methods for detecting sub-patterns in signals.

While (6) does not specify a unique method for isolating the particular subset of the column $\{i\}$ indices from $P_N$, it does offer the possibility for an alternative representation of the original problem as follows. We first consider a map from the original kernel rows/columns to a set of characteristic integer sequences of the form

$$\vec{K}_i \to \sigma_i = \sum_{j=1}^{L^2} K_{ij} 2^{j-1} \tag{7}$$

Naturally, the ones complement corresponds to $\{\bar{\sigma}_i\} = 2^{L^2} - \{\sigma_i\} - 1$. In figure 4 we show the log plot of such a sequence for the case of a *7* x *7* board. Unfortunately, no specific alternative for the expression of these sequences is known to the author apart from (7) and their significance is further discussed in the last section.

It is now possible to provide a compressed representation of (6) after introducing a pointwise or "bitwise" logical AND operator over two integers as $\vee(v,\mu)$ replacing the pointwise products with the logical formula

$$\sigma_S = \vee\left(\sigma_{p_L}, \vee .... \vee \left(\sigma_{p2}, \sigma_{p1}\right)\right) \tag{8}$$

where $\sigma_S$ denotes the original binary configuration vector also encoded via a polynomial representation as in (7). This last formula together with the original property in (6) suggests the existence of a generic function in $L$ variables acting as a "potential" for this class of problems. Moreover, there is a certain connection of (8) with a class of Boolean fractal matrices that deserves further attention.

It is a kind of 'folklore theorem' that all matrices coming from pointwise logical operations with two or more arguments have an arithmetic fractal structure. This structure becomes evident when viewed over a hierarchy of exponential intervals of integers. Justification comes from the fact that any binary word dictionary can be decomposed to a set of periodic functions with exponentially increasing period. The same exact structure appears in the well known Rademacher basis used in the construction of the Walsh functions [25] inside the unit interval. It is possible to write an explicit recursive relation for any order for the basic logical matrices as

$$T_{n+1} = \begin{bmatrix} T_n + 2^{m_{11}} & T_n + 2^{m_{12}} \\ T_n + 2^{m_{21}} & T_n + 2^{m_{22}} \end{bmatrix} \tag{9}$$

where $m$ an appropriate exponent matrix which is in fact the *2 x 2* representation of the associated two bits logical operation. Hence, for (8) we have

$$m_\vee = \begin{pmatrix} 0 & 1 \\ 1 & 1 \end{pmatrix}$$

The same can easily be given for the corresponding digit-sums directly derivable from (9) as

$$S_{n+1}^{2(T)} = \begin{bmatrix} S_n^{2(T)} & S_n^{2(T)}+1 \\ S_n^{2(T)}+1 & S_n^{2(T)}+1 \end{bmatrix} \quad (10)$$

Given the above definitions it is also possible to redefine the original problem in a hierarchy of N-dimensional hypercubes $h_N$ of successive ($\vee$) products in which case any N-queens problem becomes that of locating the unique paths associated with $s_2(h_N) = N$ at the Nth dimension when sampled at the particular coordinates given by the characteristic kernel sequences.

## 5. Discussion and conclusions

In the present, a preliminary examination of certain underlying properties of the n-queens problem is suggested via the introduction of an equivalent effectively one dimensional representation, based on certain topology encoding matrices or "kernels" for the original queen neighborhoods. An important property of these kernel rows permits a direct association with any configuration vector describing a valid solution for the original checkerboard. After further simplification, it is shown that any kernel can be associated with a unique integer sequence and any valid configuration vector is identical with a composition of certain pointwise (bitwise) logical operators acting on these sequences.

The particular sequences obtain via the explicit kernel construction do not appear to be part of any list of known ones, either in the OEIS database or elsewhere at least to the author's knowledge. Moreover, since all pointwise logical operators reveal a specific arithmetic fractal structure the identity between their functional composition over these sequences and the solution vectors may be revealing for the existence of an underlying geometry characterizing the total solutions set.

The study of any structure in these sequences for higher order problems requires special algorithms with unbounded integers which could be reported in future work. The fact that all kernel matrices form a hierarchy suggests that their study could be facilitated with a special method of the inductive combinatorial hierarchies originally introduced in [25]. Lastly, it should be mentioned that the "holistic" approach offered by the kernel method may be well suited for certain machine learning and convolution network applications for the location of the dyadic substructures associated with the solutions set.

## Appendix: Matlab code for the kernel constructor

```
function [K, Kc, Kd, newdim] = kernel(dim, arith, report)
% The overll interaction kernel is constructed via the two composites
% Kc for the 'cross' like parts of any queen's neighborhood and
% Kd for the diagonals with dim is the dimension of the original board.
% If the 2nd arg arith is present, the equivalent integer sequence is
% given in the output instead of the matrix.
if nargin<3, report=0; end
if nargin<2, arith=0; end
clc, close all
Kc = []; Kd = Kc; newdim = dim^2;
v = 0:dim-1;m1=v; % pointer masks for diagonals
for i=1:dim-1,m1 = [m1; v-i];end
m2 = dim - abs(m1);
m3 = fliplr(m1);
m4 = fliplr(m2);
for i=1:dim
    for j=1:dim
        wc = zeros(dim, dim); wd = wc;
        wc(i, :) = 1; wc(:, j) = 1;
        wd = diag( ones( m2(i, j), 1 ), m1(i, j) );
        wd = wd + fliplr( diag( ones( m4(i, j), 1), m3(i, j) ) );
        wc(i, j) = 0; wd(i, j) = 0;
        Kc = [Kc; reshape( wc', 1, newdim )];
        Kd = [Kd; reshape( wd', 1, newdim )];
```

```
        K = Kc + Kd;
      % imagesc(1-w), title(['i = ',num2str(i),' j = ',num2str(j)]), colormap gray;pause
   end
end
if arith, K = K*( 2.^( 0:newdim-1 ) )'; end
figure(1), spy(K), title('Band structure of kernel matrix')
[a b] = eig(K);
figure(2), bar( diag(b) ), title 'Eigenvalue Spectrum for kernel'
figure(3), imagesc( abs(a) ), title 'Eigenvectors for kernel'
end
end
```

**Figures**

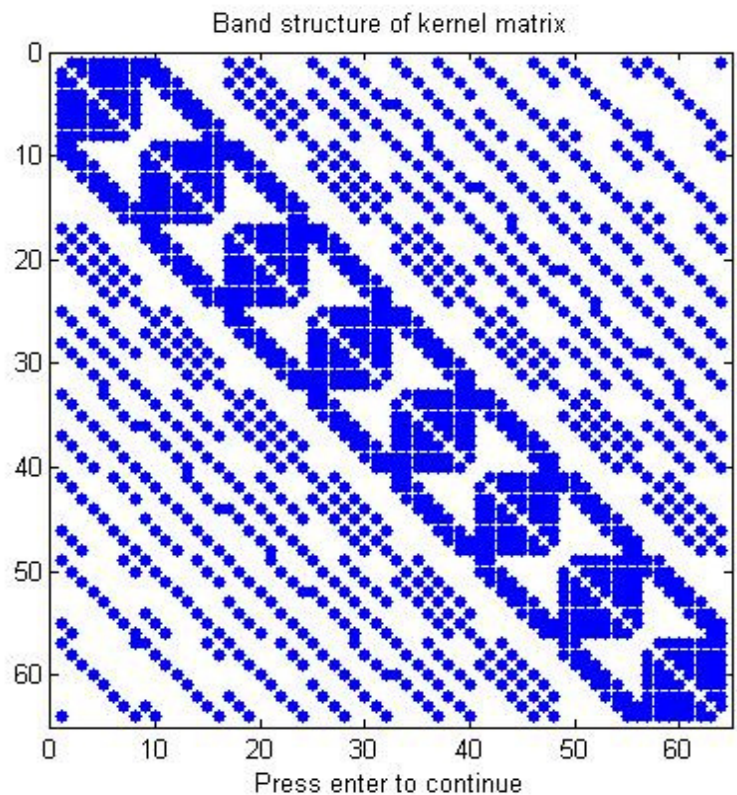

**Fig. 1** The Interaction Kernel for an 8 x 8 chessboard.

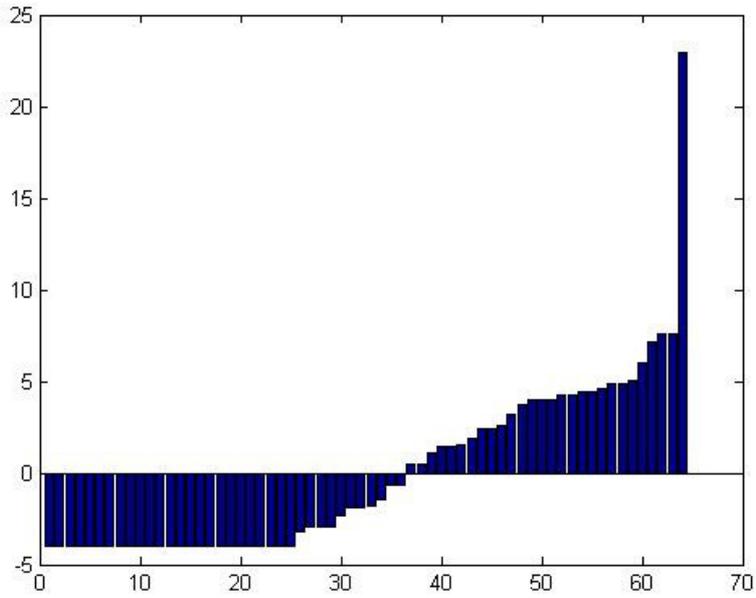

**Fig 2** The eigenvalue spectrum for the kernel of figure 1.

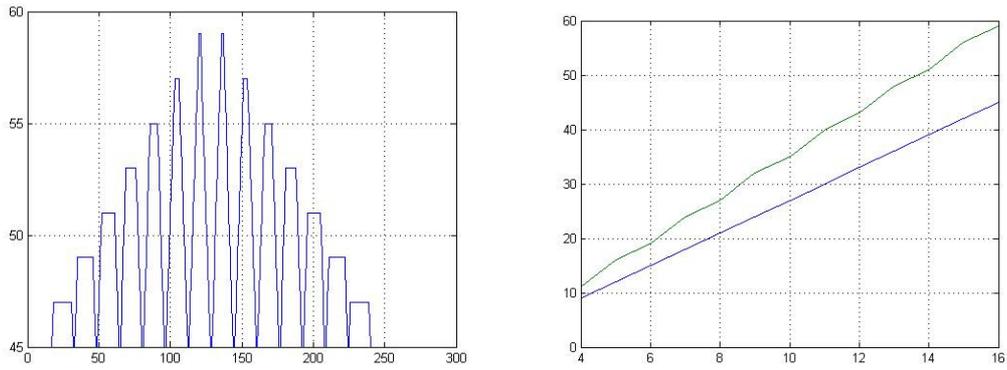

**Fig. 3** (a) Bit summands of each row of a 16 x 16 board kernel, (b) evolution of lower and upper bounds for bit summands with board size.

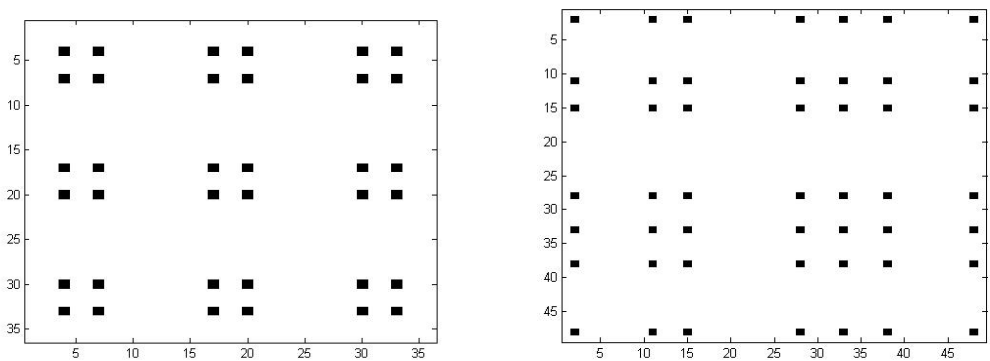

**Fig. 4** The dyadic submatrices of the kernel's complement for (a) the 6 x 6 problem and, (b) the 7 x 7 problem.

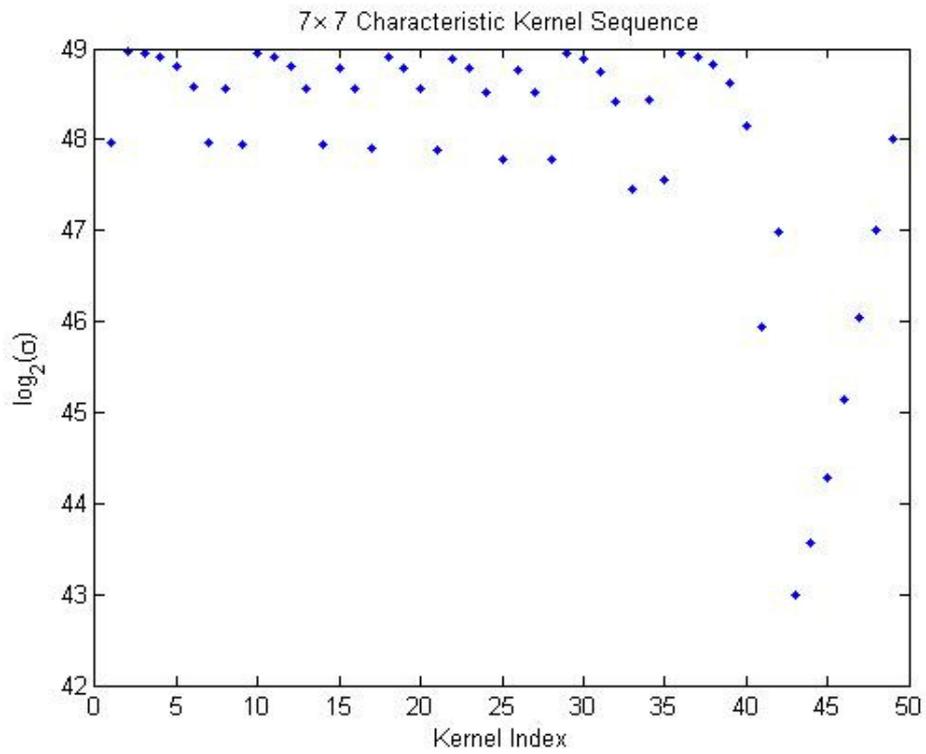

**Fig. 5** The characteristic integer sequence for the *7 x 7* chessboard kernel.